# Automating Expert-Level Medical Reasoning Evaluation of Large Language Models

Shuang Zhou[1,*], Wenya Xie[2,*], Jiaxi Li[3,*], Zaifu Zhan[4], Meijia Song[5], Han Yang[6], Cheyenna Espinoza[7], Lindsay Welton[7], Xinnie Mai[8], Yanwei Jin[9], Zidu Xu[10], Yuen-Hei Chung[11], Yiyun Xing[12], Meng-Han Tsai[13], Emma Schaffer[7], Yucheng Shi[3], Ninghao Liu[3], Zirui Liu[2], Rui Zhang[1,+]

**Affiliations:**
1. Division of Computational Health Sciences, Department of Surgery, University of Minnesota, Minneapolis, MN, USA
2. College of Science and Engineering, University of Minnesota, Minneapolis, MN, USA
3. School of Computing, University of Georgia, Athens, GA, USA
4. Department of Electrical and Computer Engineering, University of Minnesota, Minneapolis, MN, USA
5. School of Nursing, University of Minnesota, Minneapolis, MN, USA
6. Institute for Health Informatics, University of Minnesota, Minneapolis, MN, USA
7. Department of Surgery, University of Minnesota, Minneapolis, MN, USA
8. School of Data Science, University of Virginia, Charlottesville, VA, USA
9. Division of Biostatistics & Health Data Science, University of Minnesota, Minneapolis, MN, USA
10. School of Nursing, Columbia University, New York, New York, USA
11. Division of Cardiac Electrophysiology, University of California San Francisco, San Francisco, CA, USA
12. School of Dentistry, University of Minnesota, Minneapolis, Minnesota, USA
13. Division of Cardiothoracic Surgery, Department of Surgery, University of Colorado Anschutz Medical Campus, Aurora, CO, USA

*These authors contributed equally to this work.
+Correspondence: ruizhang@umn.edu

# Abstract

As large language models (LLMs) become increasingly integrated into clinical decision-making, ensuring transparent and trustworthy reasoning is essential. However, existing evaluation strategies of LLMs' medical reasoning capability either suffer from unsatisfactory assessment or poor scalability, and a rigorous benchmark remains lacking. To address this, we introduce MedThink-Bench, a benchmark designed for rigorous, explainable, and scalable assessment of LLMs' medical reasoning. MedThink-Bench comprises 500 challenging questions across ten medical domains, each annotated with expert-crafted step-by-step rationales. Building on this, we propose LLM-w-Ref, a novel evaluation framework that leverages fine-grained rationales and LLM-as-a-Judge mechanisms to assess intermediate reasoning with expert-level fidelity while maintaining scalability. Experiments show that LLM-w-Ref exhibits a strong positive correlation with expert judgments. Benchmarking twelve state-of-the-art LLMs, we find that smaller models (e.g., MedGemma-27B) can surpass larger proprietary counterparts (e.g., OpenAI-o3). Overall, MedThink-Bench offers a foundational tool for evaluating LLMs' medical reasoning, advancing their safe and responsible deployment in clinical practice.

# Introduction

Large language models (LLMs) have made remarkable progress in clinical decision-making, demonstrating the ability to perform complex reasoning tasks such as disease diagnosis[1,2], treatment planning[3], and patient management[4]. Despite their impressive capabilities, the opaque and black-box nature of LLMs limits their reliability in high-stakes clinical scenarios[5,6]. For instance, an LLM may arrive at the correct diagnosis based on parametric knowledge without providing evidence grounded in clinical guidelines or considering a comprehensive differential diagnosis[7–9]. Moreover, LLMs are prone to hallucinations, generating plausible but factually incorrect information that can mislead clinical decision-making[10–12]. Such behavior poses potential risks to patient safety and undermines the reliability of clinical workflows[13]. Therefore, deploying LLMs in clinical practice requires not only high prediction accuracy but also transparent and explainable reasoning processes[14].

Evaluating the medical reasoning capabilities of LLMs is crucial for establishing trust and ensuring safe integration into healthcare settings[15]. Recent efforts in this direction have followed two main approaches. The first approach involves assessing performance on complex medical exercises, such as multiple-choice questions (MCQs), by measuring prediction accuracy[16–18]. While this method aligns with the prevailing approaches to evaluating LLMs' medical capabilities and offers a coarse measurement[19,20], it fails to capture the depth and validity of the reasoning process that underpins clinical decisions and cannot identify flawed reasoning[21,22]. The second approach focuses on evaluating the rationales provided by LLMs, which can well address the above issues. Within this category, evaluation strategies can be further classified into three types: (1) text-similarity metrics, which compare LLM-generated rationales with reference rationales[23,24]; (2)

human expert evaluation, which relies on domain experts' manual efforts to assess the reasoning quality[14,22,25,26]; and (3) LLM-as-a-Judge, where a separate LLM is used to assess the quality of the reasoning process[27,28].

Despite these advances, existing evaluation strategies either suffer from unsatisfactory assessment or poor scalability. Specifically, while conventional text similarity metrics are scalable and cost-efficient[5], such as those based on token-level overlap (e.g., BLUE[29], ROUGE[30]) or semantic similarity (e.g., BERTScore[31]), they fail to capture medical semantics or nuanced logics and lack robustness to the variance of the text's expression styles[5]. In contrast, human evaluation remains the gold standard for assessing factuality and nuance, but is labor-intensive and limited in scale[22]. Additionally, LLM-as-a-Judge offers a scalable alternative and can comprehend medical knowledge[27,32], yet it is vulnerable to hallucinations and evaluative bias[33]. As such, the challenge of conducting automated, scalable evaluations of LLMs' medical reasoning while maintaining expert-level factuality remains unresolved.

A further barrier is the lack of a benchmark designed to evaluate LLM-generated medical rationales rigorously. Existing datasets often suffer from narrow clinical scenarios[34] or rely on LLM-generated rationales as reference answers[28,35], which may involve incorrect knowledge or flawed rationale. For example, Wu et al.[34] constructed a diagnostic benchmark using LLM-generated reasoning trajectories as ground truth. However, the credibility of such artificial intelligence (AI)-generated rationales is uncertain[8,36,37], and the alignment with human expert judgment remains unclear[38].

To address these gaps, we introduce MedThink-Bench, a benchmark tailored for rigorous, explainable, and scalable evaluation of LLMs' medical reasoning. MedThink-Bench comprises 500 challenging medical questions (QA) across ten representative domains, each annotated by medical professionals with fine-grained, step-by-step reasoning trajectories that mirror real-world clinical logic. Building on this, we propose LLM-w-Ref, an evaluation framework that integrates expert-curated fine-grained rationales with LLM-as-a-Judge mechanisms, thus combining their advantages. By calibrating the LLM-based evaluator with nuanced reasoning trajectories, our framework can accurately assess the intermediate reasoning to achieve expert-level factual consistency while enabling scalable assessments.

In this study, we demonstrate that LLM-w-Ref correlates strongly with expert evaluation and remains robust across different prompts and base models. Moreover, our benchmark comparison of twelve state-of-the-art LLMs reveals a surprising finding: smaller models such as MedGemma-27B and HuatuoGPT-70B[39] can outperform larger proprietary models like OpenAI-o3 and DeepSeek-R1 in medical reasoning. Our contributions are threefold. First, we address the longstanding challenge of scalable and expert-level evaluation of LLM-generated medical rationales. Second, we construct a high-quality dataset featuring 500 expert-annotated questions with nuanced reasoning trajectories across ten medical domains. Third, we provide a comprehensive comparison of twelve LLMs in terms of their medical reasoning capabilities. Overall, MedThink-Bench offers a foundational resource for assessing the trustworthiness of

LLMs in medical decision-making, thereby advancing their safe and responsible integration into clinical practice.

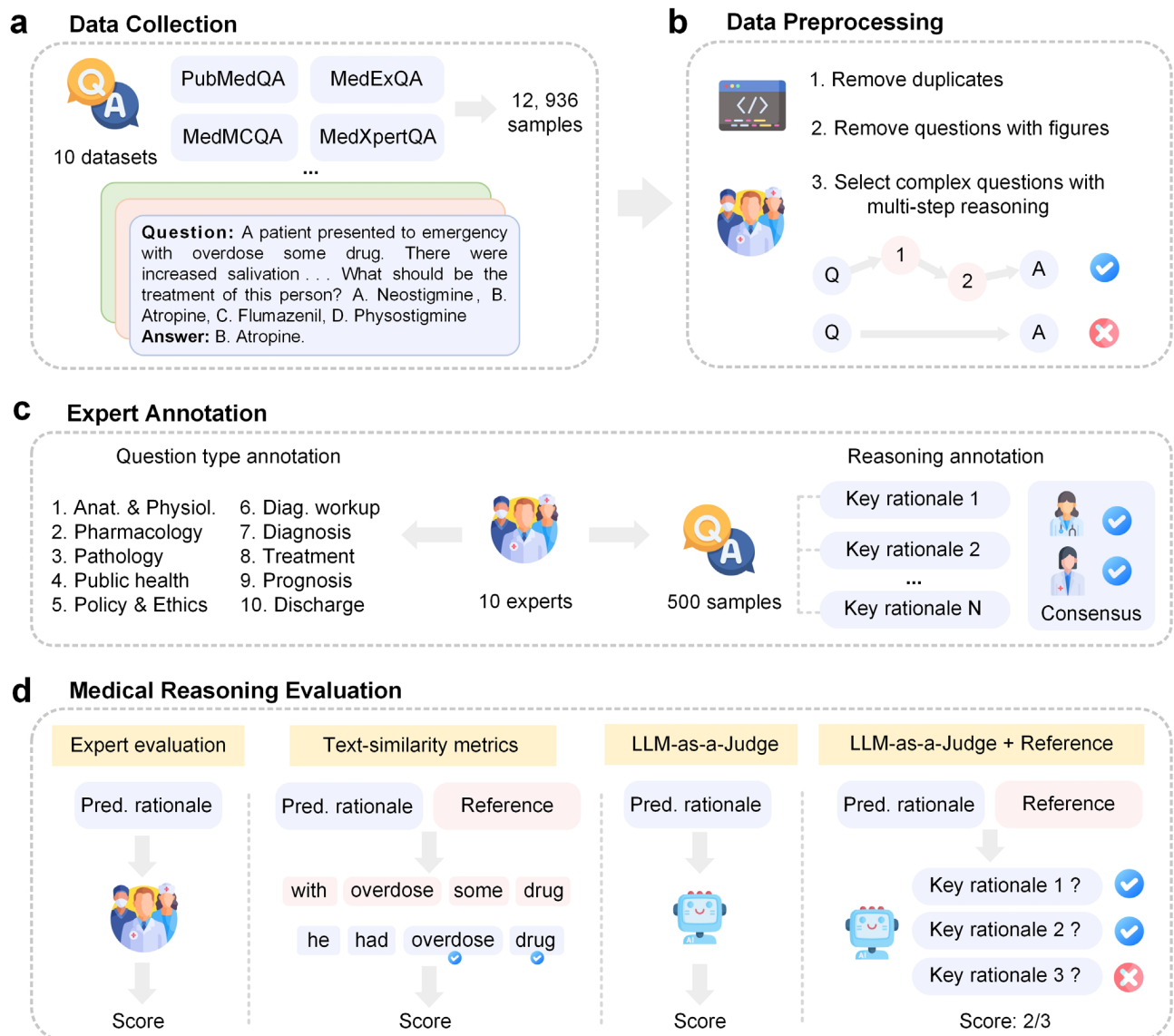

Figure 1. Overview of the MedThink-Bench dataset. a. Data collection. Medical questions were sourced from ten publicly available datasets, each accompanied by ground-truth answers. b. Data preprocessing. Duplicate entries and questions involving medical images were removed. Medical experts then manually curated a subset of complex questions requiring multi-step reasoning. c. Expert annotation. A team of ten medical experts annotated the questions into ten distinct medical domains and collaboratively generated fine-grained reasoning trajectories through consensus. d. Medical reasoning evaluation. We rigorously evaluated the medical reasoning capabilities of twelve LLMs, comparing them against expert evaluations, text-similarity metrics, LLM-as-a-Judge, and reference-based LLM-as-a-Judge (LLM-w-Ref). Additionally, we analyzed the correlation between these automated metrics and expert evaluations.

# Results

In this section, we present the key findings of our study. We first provide an overview of the MedThink-Bench dataset, followed by a comprehensive comparison of mainstream metrics for evaluating medical reasoning in twelve LLMs and robustness analyses of the LLM-w-Ref framework.

## Dataset

We created MedThink-Bench, a medical QA dataset with expert-derived reasoning annotations, comprising 500 complex questions across ten medical scenarios. The raw data were sourced from publicly available medical QA datasets (Supplementary Data 1). The ten domains covered are: Pathology, Discharge, Disease Diagnosis, Anatomy & Physiology, Treatment, Public Health, Policy & Ethics, Prognosis, Diagnostic Workup, and Pharmacology. Our annotation team manually selected complex questions that require multi-step reasoning and provided fine-grained reasoning step annotations. The annotation procedure is described in Supplementary Data 2. The dataset statistics are shown in Fig. 2.

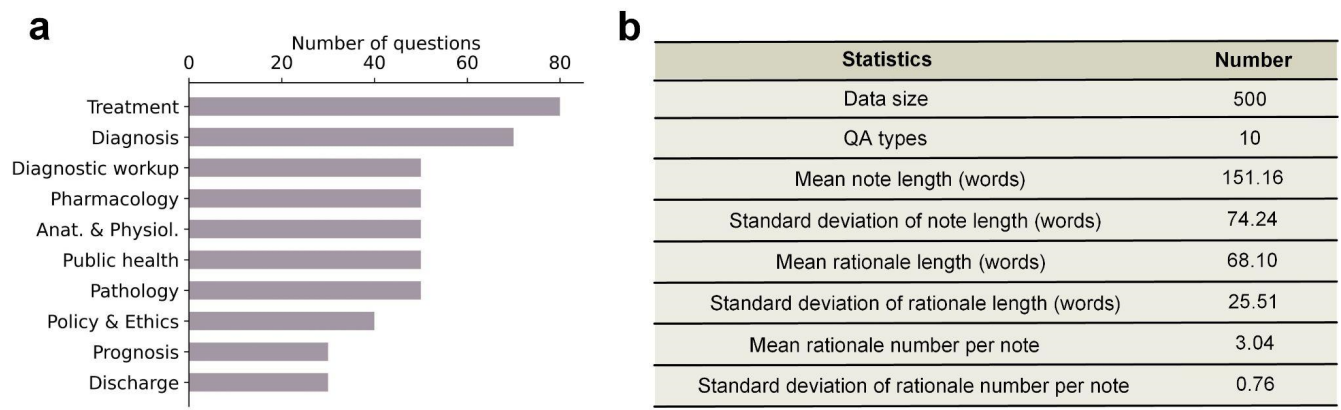

Figure 2. Dataset statistics of MedThink-Bench. a. Breakdown of the ten medical domains included in the MedThink-Bench dataset. b. Detailed statistics of the dataset.

## Comparison of evaluation metrics on LLM reasoning

We evaluated the reasoning performance of twelve LLMs on the MedThink-Bench dataset, focusing on medical reasoning abilities and comparing mainstream evaluation metrics. We assessed the reasoning performance of twelve LLMs with zero-shot Chain-of-Thought (CoT) prompting[40], as described in the Methods section. The comparison included both expert evaluations and eight automated metrics, with the results presented in Fig. 2. Additionally, we compared the prediction accuracy of the LLMs, with detailed results available in Supplementary Data 3. As shown in Fig. 3, the expert evaluation scores range from 0.453 (95% confidence interval (CI): 0.419–0.485) for Med42-70B to 0.759 (95% CI: 0.730–0.789) for MedGemma-27B. The performance of reference-based LLM-as-a-Judge (LLM-w-Ref) varies from 0.482 (95% CI:

0.450–0.514) for Med42-70B to 0.769 (95% CI: 0.742–0.798) for MedGemma-27B. Meanwhile, the performance of reference-free LLM-as-a-Judge (LLM-w/o-Ref) ranges from 0.823 (95% CI: 0.812–0.834) for Med42-70B to 0.907 (95% CI: 0.896–0.918) for Qwen3-32B. Among the text-similarity metrics, BLUERT and BERTScore generally outperformed the other metrics. Specifically, BLUERT scores range from 0.395 (95% CI: 0.388–0.403) for OpenAI-o3 to 0.599 (95% CI: 0.589–0.608) for MedGemma-27B, while BERTScore ranges from 0.554 (95% CI: 0.551–0.557) for HuatuoGPT-o1-70B to 0.630 (95% CI: 0.625–0.635) for Med42-70B.

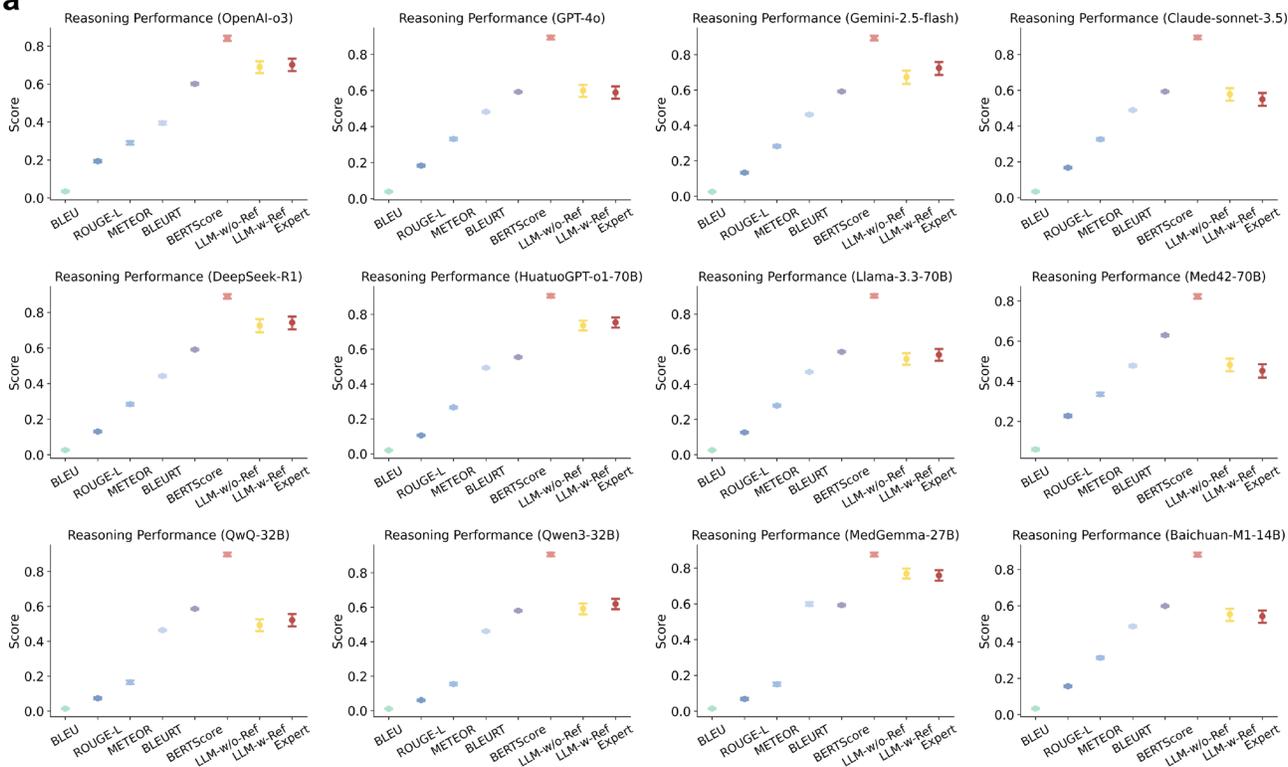

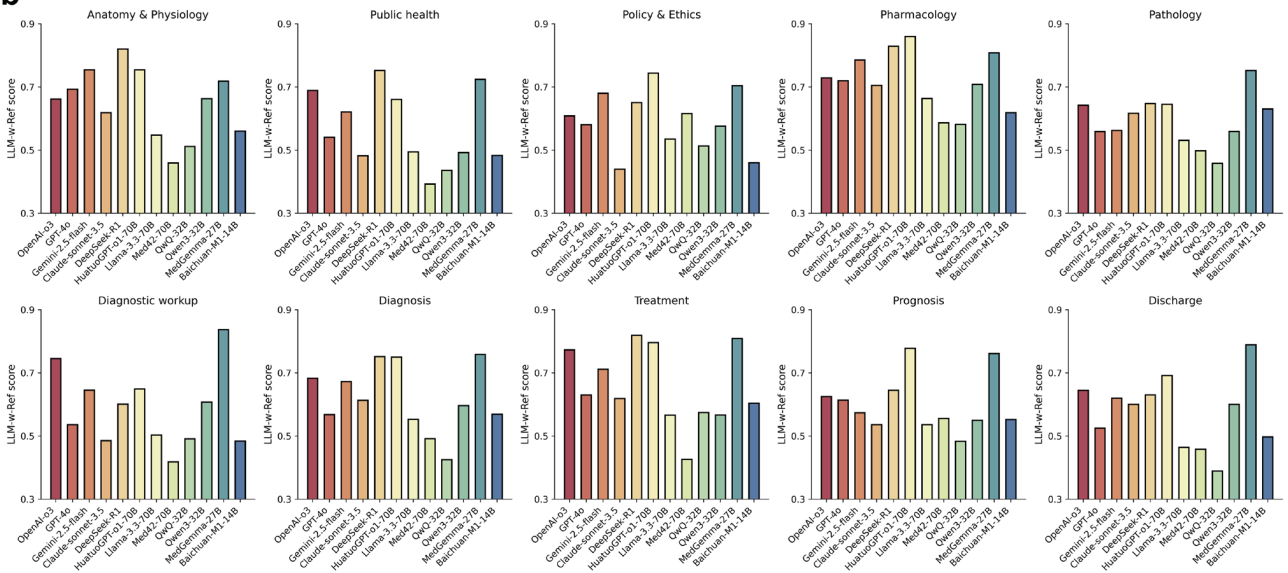

Figure 3. Medical reasoning performance on the MedThink-Bench dataset. a. Comparison of overall medical reasoning performance, including expert evaluation, five text-similarity metrics, and the proposed LLM-w-Ref framework under zero-shot prompting. The automated reasoning assessments were obtained by comparing ground-truth reasoning annotations with the predicted annotations. Error bars represent the 95% CI of the mean, calculated via bootstrapping. b. Breakdown of medical reasoning performance across the ten medical domains in the MedThink-Bench dataset.

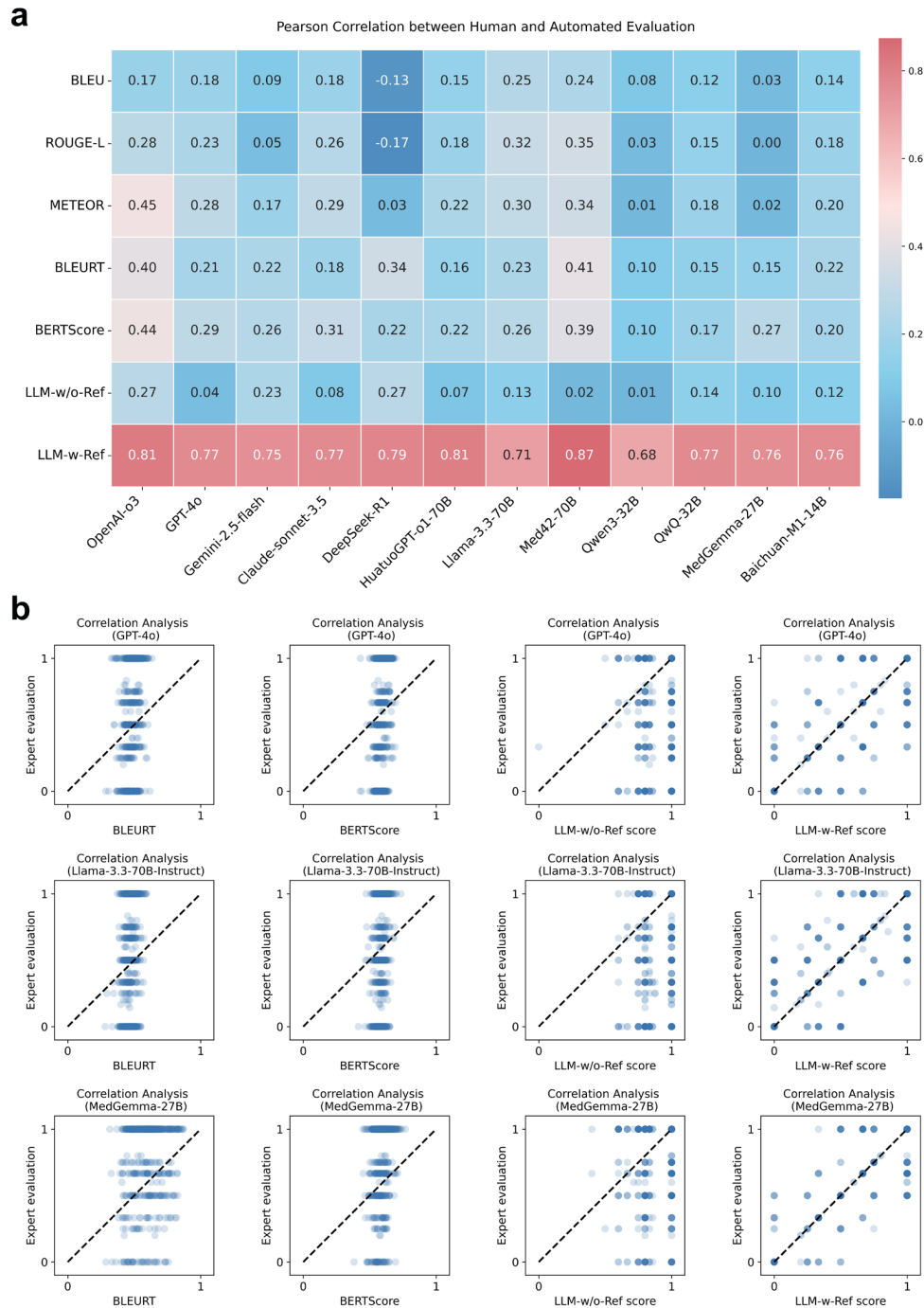

Figure 4. Correlation analysis between expert and automated evaluation of medical reasoning performance. a. Pearson correlation analysis of predicted rationales from Llama-3.3-70B, evaluated against expert assessments and various automated metrics. These metrics include text-similarity measures (BLEU, ROUGE-L, METEOR, BLEURT, BERTScore), LLM-w/o-Ref (which does not use ground-truth rationales as a reference), and LLM-w-Ref (which uses our annotated fine-grained rationales as a reference). The results indicate a strong correlation between LLM-w-Ref and expert evaluations, while LLM-w/o-Ref and text-based metrics show weaker correlations with expert assessments. b.Scatter plot showing the correlation between expert evaluation scores and automated scores for each sample in the MedThink-Bench dataset. The plot includes results for GPT-4o, Llama-3.3-70B, and MedGemma-27B. Each point represents an individual sample, with dashed lines indicating equal performance between expert and automated scores. For LLM-w-Ref, many data points align closely with the dashed line, suggesting strong agreement with expert evaluations. In contrast, BLEURT, BERTScore, and LLM-w/o-Ref exhibit greater divergence from the dashed line, indicating a weaker alignment with expert assessments.

## Correlation analysis between expert and automated evaluation

We computed Pearson correlation coefficients between the expert evaluation scores and all the automated metrics across all LLMs, as shown in Fig. 4a. The results reveal weak correlations for metrics such as BLEU, ROUGE-L, METEOR, BLEURT, and BERTScore, with Pearson coefficients ranging from -0.17 to 0.45. Similarly, LLM-w/o-Ref showed weak correlation, with coefficients ranging from 0.01 to 0.27. In contrast, LLM-w-Ref demonstrated a strong correlation with expert evaluations, with Pearson coefficients ranging from 0.68 to 0.87. We further visualized the individual evaluation scores of Llama-3.3-70B in Fig. 4b and Supplementary Data 4. Data points for LLM-w-Ref generally closely align with the dashed line, indicating agreement with expert scores. In contrast, BLEURT, BERTScore, and LLM-w/o-Ref show greater divergence from expert evaluations.

## The impact of the judge model in LLM-w-Ref

We assessed the robustness of the LLM-w-Ref framework by testing it with various LLMs as the judge model. Using the predicted rationales from Llama-3.3-70B, we evaluated the performance of ten LLMs of different scales, including GPT-4o-mini, DeepSeek-R1, Gemini-2.5-flash, HuatuoGPT-o1-70B, Qwen3-32B, QwQ-32B, MedGemma-27B, Llama-3-8B-Instruct, Llama-3.2-3B-Instruct, and Llama-3.2-1B-Instruct. The results, presented in Fig. 5a, show reasoning performance ranging from 0.522 (95% CI: 0.492–0.550) for Gemini-2.5-flash to 0.553 (95% CI: 0.515–0.587) for HuatuoGPT-o1-70B. However, when using smaller models as the judge models, such as Llama-3-8B-Instruct, Llama-3.2-3B-Instruct, and Llama-3.2-1B-Instruct, the performance is higher than that of the large models, which range from 0.704 to 0.891.

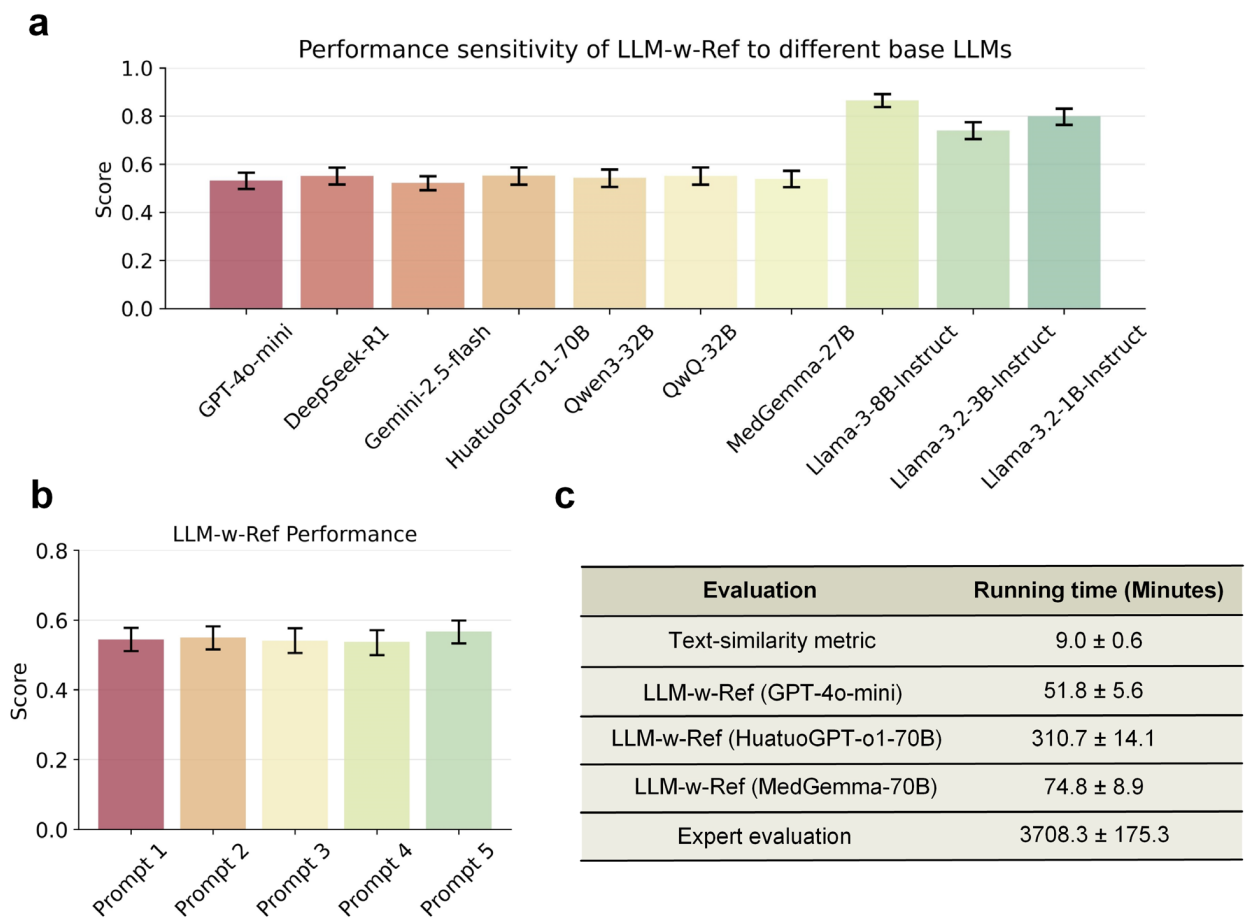

Figure 5. Performance and efficiency analysis of LLM-w-Ref. a. Robustness of LLM-w-Ref across different judge models. Predicted rationales from Llama-3.3-70B were evaluated using ten LLMs of varying scales as judge models. b. Sensitivity of LLM-w-Ref to prompt variations. Five semantically similar prompt variations were tested to assess the framework's robustness. c. Efficiency comparison of rationale assessment among text-similarity metrics, LLM-w-Ref, and human evaluation. Running times for all evaluated LLMs on the MedThink-Bench dataset were recorded. The running time reported for text-similarity metrics represents the cumulative time across all metrics.

## The prompt sensitivity in LLM-w-Ref

We also investigated the sensitivity of the LLM-w-Ref framework to different prompt formulations by composing five prompt variations with similar semantics (Supplementary Data 4). The reasoning performance of each variation was compared using the predicted rationales from Llama-3.3-70B, with GPT-4o-mini fixed as the judge model. As shown in Fig. 5b, the performance varied from 0.538 (95% CI: 0.500–0.571) for the second prompt variant to 0.567 (95% CI: 0.533–0.599) for the fifth prompt.

**Type:** Disease diagnosis
**Question:** A 62-year-old patient has been hospitalized for a week due to a stroke. One week into the hospitalization, he develops a fever and purulent cough. His vitals include: heart rate 88/min, respiratory rate 20/min, temperature 38.4°C (101.1°F), and blood pressure 110/85 mm Hg. On physical examination, he has basal crackles on the right side of the chest. Chest radiography shows a new consolidation on the same side. Complete blood count is as follows: Hemoglobin 16 mg/dL, Hematocrit 50%, Leukocyte count 8,900/mm3, Neutrophils 72%, Bands 4%, Eosinophils 2%, Basophils 0%, Lymphocytes 17%, Monocytes 5%, Platelet count 280,000/mm3. What is the most likely causal microorganism?
A. Pseudomona aeruginosa, B. Streptococcus pneumoniae, C. Mycobacterium tuberculosis, D. Haemophilus influenzae, E. Staphylococcus aureus.

**Ground-truth answer:** E. Staphylococcus aureus.
**Predicted answer:** B. Streptococcus pneumoniae.

**Fine-grained reasoning annotation:**
- Laboratory Results: leukocyte count 8,900/mm^3, Neutrophils 72%, together with the high body temperature 38.4°C (101.1°F), suggests a bacterial infection.
- A hospitalized patient who develops new fever, purulent cough, and consolidation on chest X-ray one week after admission is suspicious for hospital-acquired pneumonia (HAP).
- For non-ventilated early-onset HAP patients, Staphylococcus aureus is the most common pathogen.

**Predicted rationale:**
**Symptoms**: The patient develops a fever, purulent cough, and has basal crackles on the right side of the chest, with a new consolidation seen on chest radiography. These symptoms are indicative of pneumonia.
**Vitals and Laboratory Results**: The patient's temperature is elevated at 38.4°C (101.1°F), indicating an infection. The leukocyte count is slightly elevated at 8,900/mm^3, with a predominance of neutrophils (72%), which suggests a bacterial infection.
Given the patient's hospital setting, the acute onset of symptoms, and the presence of purulent cough and fever, the most likely causal microorganisms would be those commonly associated with hospital-acquired pneumonia.
Streptococcus pneumoniae can cause hospital-acquired pneumonia, especially in patients who have not been previously exposed to broad-spectrum antibiotics. The final answer is B.

**Reasoning correctness: 2/3**

Figure 6. Case study of medical reasoning assessment. This case demonstrates that while the prediction model Llama-3.3-70B produced an incorrect answer, it followed partially correct medical reasoning trajectories. This highlights the advantage of the LLM-w-Ref framework, which, in conjunction with the fine-grained rationale annotations in MedThink-Bench, provides a more nuanced evaluation of the medical reasoning abilities of LLMs compared to merely assessing prediction accuracy.

## Efficiency comparison

We compared the time efficiency of various evaluation strategies to assess their scalability. Specifically, we recorded the running times for human evaluation, text-similarity metrics, and LLM-w-Ref evaluations on MedThink-Bench. As shown in Fig. 5c, the average assessment time for text-similarity metrics was 9.0 minutes, while LLM-w-Ref took 310.7 minutes when using

HuatuoGPT-o1-70B as the judge model. In contrast, human evaluation required an average of 3708.3 minutes, significantly longer than the automated methods.

## Case study

Case studies were conducted to demonstrate the effectiveness of LLM-w-Ref in measuring the medical reasoning capability of LLMs. As shown in Fig. 6, Llama-3.3-70B produced an incorrect answer, yet followed a partially correct medical reasoning trajectory. In another case (Supplementary Data 5), flawed reasoning led to a correct answer. By measuring medical reasoning with expert-curated fine-grained rationales, LLM-w-Ref captures flawed reasoning patterns and offers a more nuanced evaluation of an LLM's medical capabilities. These findings underscore the importance of assessing not only the final prediction but also the underlying reasoning process.

# Discussion

In this study, we present MedThink-Bench, a curated dataset comprising complex medical questions spanning ten clinical scenarios, each accompanied by fine-grained rationale annotations from domain experts. MedThink-Bench is designed to address a critical challenge in evaluating LLMs for medical reasoning: enabling automated assessment while maintaining expert-level factual accuracy. Below, we summarize the key findings and insights derived from our study.

First, MedThink-Bench effectively differentiates the reasoning capabilities of the LLMs (Fig. 3). Our benchmarking results show that MedGemma-27B, HuatuoGPT-o1-70B, and DeepSeek-R1 are the top three performers in medical reasoning, with scores of 0.769 (95% CI: 0.742–0.798), 0.737 (95% CI: 0.708–0.764), and 0.727 (95% CI: 0.689–0.762), respectively. Notably, smaller open-source models such as MedGemma-27B outperformed larger commercial counterparts, including OpenAI-o3 and DeepSeek-R1. Similarly, Qwen3-32B significantly outperformed Gemini-2.5-flash and DeepSeek-R1 ($p < 0.001$). Among commercial LLMs, DeepSeek-R1, OpenAI-o3, and Gemini-2.5-flash emerged as the top performers, substantially surpassing GPT-4o and Claude-sonnet-3.5.

Second, we introduce an evaluation strategy, LLM-w-Ref, which combines LLM-as-a-Judge with fine-grained rationale annotations from MedThink-Bench. This approach exhibits strong alignment with expert evaluations (Fig. 4a). As shown in Fig. 4a, the Pearson correlation coefficients between LLM-w-Ref and expert scores range from 0.68 to 0.87, indicating a consistently strong relationship. Visualization of the score distributions (Fig. 4b) further confirms this correlation. Importantly, LLM-w-Ref demonstrates significantly higher efficiency compared to human evaluation (Fig. 5c), making it well-suited for scalable, automated evaluation. Together, these results demonstrate that integrating LLM-as-a-Judge with fine-grained rationales enables efficient, reliable, and expert-aligned evaluation of medical reasoning.

Third, our analysis reveals that conventional reasoning evaluation methods, including text alignment metrics and LLM-w/o-Ref, have a weak correlation with expert judgments (Fig. 4a). This is largely because text alignment metrics, such as BLEU or ROUGE, rely on surface-level word overlap and fail to capture semantic or logical equivalence. Even advanced metrics like BERTScore, which incorporate word embeddings, are limited in two respects: (1) they measure similarity at the word level rather than at the level of sentences or reasoning chains, and (2) they cannot comprehend the underlying logical structure of complex medical justifications. Consequently, these metrics tend to produce scores within a narrow range, as LLMs often repeat contextual information or generate plausible-sounding rationales without truly accurate reasoning (Fig. 4b). In contrast, LLM-w/o-Ref lacks access to ground-truth rationales and is highly dependent on the evaluator model's performance and biases, rendering its assessment unreliable. Overall, these findings underscore the limitations of existing strategies for evaluating medical reasoning.

We further demonstrate that LLM-w-Ref is robust to variations in both the judge model and prompt phrasing. As shown in Fig. 5a, the evaluation performance remains stable when using LLMs with strong instruction-following capabilities, such as GPT-4o-mini or MedGemma-27B, as judge models. Likewise, results in Fig. 5b indicate that prompt variants with similar semantics yield consistent evaluation outcomes, suggesting robustness to prompt engineering. This stability enhances the practical applicability of LLM-w-Ref for widespread use in reasoning evaluation.

An additional finding is that LLM reasoning performance does not consistently align with MCQ prediction accuracy (Supplementary Data 3). For example, although OpenAI-o3 trails behind MedGemma-27B and HuatuoGPT-o1-70B in reasoning (Fig. 3), it achieves the highest prediction accuracy at 0.692 (95% CI: 0.652–0.732), compared to 0.384 (95% CI: 0.342–0.426) and 0.49 (95% CI: 0.446–0.532), respectively. This discrepancy can be attributed to two factors. First, even when an LLM produces an incorrect answer, its accompanying rationale may still reflect partial correctness due to its extensive medical training. Such nuance is captured by rationale-based evaluation but not by prediction accuracy. Second, LLMs may occasionally arrive at correct answers through flawed reasoning, as reported in prior studies. In such cases, prediction accuracy is inflated despite poor reasoning quality. These observations collectively suggest that prediction accuracy alone is insufficient to reflect true medical reasoning ability, whereas rationale-based evaluation offers deeper insights.

Despite these advances, our study has several limitations. First, since the medical questions were derived from existing literature, there is a possibility that some LLMs may have been exposed to the corresponding answers during pretraining. Although the ground-truth rationales were newly annotated by domain experts, this potential data leakage could inflate the prediction accuracy[41] and contribute to the observed discrepancy between answer accuracy and reasoning quality. Second, due to the high cost of expert annotation, the size of the MedThink-Bench dataset is relatively limited compared to datasets constructed using AI-generated questions or rationales[34,42]. As a result, while MedThink-Bench is well-suited for rigorous model evaluation, its current scale constrains its utility for model training or fine-tuning aimed at enhancing LLMs' medical reasoning capabilities.

In summary, this study addresses the critical challenge of accurately and efficiently evaluating the medical reasoning capabilities of LLMs. To this end, we curated MedThink-Bench, a medical QA dataset covering ten clinical scenarios with fine-grained rationale annotations from domain experts. We introduced an evaluation framework, LLM-w-Ref, that combines the nuanced annotations with the LLM-as-a-Judge paradigm to enable automated yet expert-aligned reasoning assessment. Using this framework, we benchmarked the reasoning performance of twelve LLMs and systematically compared multiple evaluation strategies. Our findings show that LLM-w-Ref achieves strong concordance with expert assessments while offering substantially greater efficiency. Overall, this work provides a robust and scalable solution for evaluating LLMs' medical reasoning, reducing the burden of manual evaluation and paving the way for their integration into clinical practice.

# Methods

## Data curation

We collected comprehensive medical questions from ten existing datasets, filtered out hard questions that required multi-step reasoning (Supplementary Data 2), and annotated the reasoning rationales by experts. Specifically, the medical QA datasets from which we collected questions included MedBullets[43], MMLU-Pro[44], MedExQA[45], MedXpertQA[18], Humanity's Last Exam[46], MedQA-USMLE[47], PubMedQA[48], MedMCQA[49], MMLU-Medicine[50], and HEAD-QA[51]. The statistics of these datasets are shown in Fig. 2. Then, we pre-processed the collected data by removing duplicates and filtering out the questions involving images for prediction.

## Data annotation

We built a well-annotated dataset to facilitate automated expert-level reasoning evaluation. The medical questions were divided into ten medical domains: Pathology, Discharge, Disease Diagnosis, Anatomy & Physiology, Treatment, Public Health, Policy & Ethics, Prognosis, Diagnostic Workup, and Pharmacology. We employed ten medical experts to curate the dataset manually. Two independent physicians annotated each medical question. When disagreement existed in the annotation, a third physician examined the case and made the final annotation. We checked the inter-annotator agreement (IAA) on the reasoning and question types (Supplementary Data 2).

## Evaluation framework

To evaluate medical rationales, we apply three evaluation strategies: human evaluation, LLM-as-a-Judge, and text-similarity metrics. Notably, we proposed an evaluation framework, LLM-w-Ref, that enabled scalable, step-level assessment of clinical reasoning abilities of LLMs. Built upon

MedThink-Bench's nuanced rationales, LLM-w-Ref compared model-generated rationales against expert-annotated reasoning trajectories to evaluate their logical correctness and stepwise completeness. Additionally, the prediction accuracy is also computed to compare LLMs' medical capability. The following describes the implementation and evaluation protocol in detail.

**Human evaluation.** To validate the reliability of our evaluation framework, we conduct a human evaluation and compare the results with automatic metrics. For each question, a domain expert is provided with the medical question and the model-generated rationale. The experts then examine the correctness of the model-generated rationale. The instance-level reasoning score is defined as:

$$R^{(i)} = \frac{ExpertCovered(r_{model}^{(i)}, q^{(i)})}{ExpertRequired(q^{(i)})}.$$

Here, $r_{model}^{(i)}$ is the model-generated rationale for question $i$; $ExpertRequired(q^{(i)})$ i is the number of reasoning steps the expert deems necessary, $ExpertCovered(r_{model}^{(i)}, q^{(i)})$ is the number of those steps that the expert judges to be present in the rationale.

**Reference-based LLM-as-a-Judge.** To enable fine-grained, scalable evaluation aligned with expert reasoning, we adopt a reference-based LLM-as-a-Judge approach (LLM-w-Ref) that operates at the step level. For each question, the expert-annotated rationale comprises a list of discrete reasoning steps, provided by medical experts during dataset construction. To evaluate step-level correctness, the judge model is given the question, the model-generated rationale, and each discrete reasoning step, and is prompted to determine whether the rationale supports that step. The final reasoning score is computed as the proportion of expert steps that are successfully identified. The instance-level reasoning score is computed as:

$$R^{(i)} = \frac{|\{s \in S_{expert}^{(i)} \mid LLMJudge(s, r_{model}^{(i)}, q^{(i)}) = Yes\}|}{|S_{expert}^{(i)}|}.$$

Here, $S_{expert}^{(i)}$ is the set of expert-annotated reasoning steps for the question $i$, and $r_{model}^{(i)}$ is the model-generated rationale, as previously defined.

**Reference-free LLM-as-a-Judge.** To examine the necessity of expert-annotated rationales for accurate evaluation, we implement a commonly adopted baseline: LLM-as-a-Judge without reference. This setting allows us to compare grounded and ungrounded evaluations and to better understand the role of reference rationales in ensuring the reliable assessment of clinical reasoning. Specifically, in this setting, the judge model was provided only with the medical question and the model-generated rationale. The judge model was first asked to estimate the number of reasoning steps required to answer the question, and then to determine how many of those steps are sufficiently supported by the model-generated rationale. The instance-level score is calculated as:

$$R^{(i)} = \frac{LLMCovered(r_{model}^{(i)}, q^{(i)})}{LLMRequired(q^{(i)})}.$$

$LLMRequired(q^{(i)})$ denotes the number of reasoning steps the LLM estimates are necessary to answer the question $i$, $LLMCovered(r^{(i)}_{model}, q^{(i)})$ indicates how many of those steps the LLM judges to be sufficiently reflected in the rationale.

**Text-similarity metric.** To contextualize our evaluation framework, we report baseline performance using widely adopted metrics, including BLEU[29], ROUGE-L[30], METEOR[52], BLEURT[53], and BERTScore[31]. Metrics such as BLEU and ROUGE-L rely on surface-level token overlap, using n-gram precision or longest common subsequence to quantify similarity. While computationally efficient, they are insensitive to paraphrasing and semantic equivalence. More recent metrics like METEOR, BLEURT, and BERTScore incorporate semantic information through synonym matching, pretrained language models, or human-annotated supervision. Although these approaches better capture general linguistic similarity, they remain limited in evaluating factual accuracy, logical soundness, and clinical relevance. While these metrics offer convenient and scalable evaluation, they do not explicitly assess the logical structure, clinical validity, or step-by-step correctness of medical reasoning. Therefore, they are used here primarily for baseline comparison.

**Reasoning score aggregation.** For all step-level evaluation methods described above, we report both instance-level and dataset-level scores. Given instance-level reasoning scores $R(i)$ for each question, $i = 1, \ldots, N$, the final dataset-level score is computed as the average:

$$R = \frac{1}{N} \sum_{i=1}^{N} R^{(i)}.$$

**Prediction accuracy.** Apart from reasoning evaluation, we follow related studies and evaluate the performance of the predicted answer with accuracy using an exact match. Formally, the instance-level accuracy is:

$$A^{(i)} = 1(a_{pred}^{(i)} = a_{gold}^{(i)}),$$

where $a_{pred}^{(i)}$ and $a_{gold}^{(i)}$ denote the predicted and golden answer for the $i$-th question, the indicator function $1(\cdot)$ returns 1 if the condition is satisfied, and 0 otherwise. The final accuracy is the average over the dataset:

$$A = \frac{1}{N} \sum_{i=1}^{N} A^{(i)}.$$

**Evaluation setting.** For each sample in MedThink-Bench, the model-generated rationale and final answer are extracted using rule-based regular expressions. For human evaluation, ten medical professionals assess whether each annotated reasoning step is present in the generated rationale. For LLM-w-Ref, we use GPT-4o-mini under zero-shot prompting with CoT formatting. The model receives the medical question, generated rationale, and individual expert steps. Decoding is performed with a temperature of 0.1, a commonly adopted setting in LLM-as-a-Judge evaluations[54], using a maximum decoding length of 4096 tokens and a fixed random seed (42) for reproducibility. In the reference-free setting, all parameters remain the same. For text-similarity metrics, BLEU, ROUGE-L, METEOR, BERTScore, and BLEURT are computed using standard toolkits with default settings.

## LLM baselines

**Large language models.** We benchmark representative LLMs from both commercial closed-source LLMs and open-source LLMs. Since this work aims to assess the medical reasoning capabilities of LLMs, we include both reasoning and non-reasoning LLMs from each category. Specifically, the commercial closed-source LLMs include GPT-4o[55,56], o3[57], Claude-3.5-sonnet[58], Gemini-2.5-flash[59], DeepSeek-R1[60], and open-source LLMs include Baichuan-M1-14B[61], HuatuoGPT-o1-70B[62], MedGemma-27B[63], Llama3-Med42-70B[64,65], Llama-3.3-70B[66], Qwen3-32B[67], and QwQ-32B[67]. In the experiment, we get responses from closed-source LLMs through APIs. For open-source LLMs, we use the package vLLM to load them for faster inference.

**Parameters for LLM Inference.** The LLM inference configuration employs carefully selected hyperparameters to ensure optimal performance and reproducibility. The temperature parameter is set to 0, implementing deterministic sampling to eliminate randomness in token selection and ensure consistent outputs across multiple runs. A fixed random seed of 42 is specified to guarantee reproducible results, which is essential for scientific rigor and experimental validation. The maximum token limit is configured to 4096, providing sufficient generation capacity for comprehensive reasoning. These parameter settings collectively establish a controlled inference environment that prioritizes consistency and reliability over creative variation, aligning with the systematic evaluation requirements of our benchmarking framework.

## Data Availability

The medical questions are sourced from public datasets, which may be available for research purposes upon reasonable request.
1. Medbullets: https://huggingface.co/datasets/LangAGI-Lab/medbullets,
2. MMLU-Pro: https://huggingface.co/datasets/TIGER-Lab/MMLU-Pro,
3. MedExQA: https://huggingface.co/datasets/bluesky333/MedExQA,
4. MedXpertQA: https://huggingface.co/datasets/TsinghuaC3I/MedXpertQA,
5. Humanity's Last Exam: https://github.com/centerforaisafety/hle,
6. MedQA-USMLE: https://huggingface.co/datasets/bigbio/med_qa,
7. PubMedQA: https://huggingface.co/datasets/bigbio/pubmed_qa,
8. MedMCQA: https://huggingface.co/datasets/lighteval/med_mcqa,
9. MMLU-Medicne: https://huggingface.co/datasets/cais/mmlu,
10. HEAD-QA: https://huggingface.co/datasets/dvilares/head_qa.

The expert-curated data with rationale annotation will be released upon the acceptance of this paper.

## Code Availability

The codes will be publicly available at https://github.com/plusnli/MedThink-Bench.

## Author Contributions

## Acknowledgment

This work was supported by the National Institutes of Health's National Center for Complementary and Integrative Health under grant number R01AT009457, National Institute on Aging under grant number R01AG078154, and National Cancer Institute under grant number R01CA287413. The content is solely the responsibility of the authors and does not represent the official views of the National Institutes of Health. We also acknowledge the support from the Center for Learning Health System Sciences.

## Competing Interests

The authors declare no competing interests.